\documentclass[letterpaper]{article} 
\usepackage{aaai2026}  
\usepackage{times}  
\usepackage{helvet}  
\usepackage{courier}  
\usepackage[hyphens]{url}  
\usepackage{graphicx} 
\usepackage{natbib}  
\usepackage{caption} 
\usepackage{algorithm}
\usepackage{algorithmic}

\usepackage{amsmath}
\usepackage{amsthm}
\newtheorem{definition}{Definition}
\newtheorem{example}{Example}
\usepackage{amsfonts}
\usepackage{amssymb}
\usepackage{booktabs} 
\usepackage{multirow}
\usepackage{threeparttable}

\usepackage{newfloat}
\usepackage{listings}
\DeclareCaptionStyle{ruled}{labelfont=normalfont,labelsep=colon,strut=off} 
\floatstyle{ruled}
\newfloat{listing}{tb}{lst}{}
\floatname{listing}{Listing}
\title{LLM-Guided Quantified SMT Solving over Uninterpreted Functions}
\author {
    Kunhang Lv\textsuperscript{\rm 1,\rm 3},
    Yuhang Dong\textsuperscript{\rm 2,\rm 3},
    Rui Han\textsuperscript{\rm 1,\rm 3},
    Fuqi Jia\textsuperscript{\rm 1,\rm 3},
    Feifei Ma\textsuperscript{\rm 2,\rm 3}\thanks{Corresponding authors.},
    Jian Zhang\textsuperscript{\rm 1,\rm 3}\footnotemark[1]
}
\affiliations {
    \textsuperscript{\rm 1}SKLCS and Key Laboratory of System Software, ISCAS, Beijing, China\\
    \textsuperscript{\rm 2}Laboratory of Parallel Software and Computational Science, ISCAS, Beijing, China\\
    \textsuperscript{\rm 3}University of Chinese Academy of Sciences, Beijing, China\\
    \{lvkh, maff, zj\}@ios.ac.cn
}

\usepackage{bibentry}

\begin{document}
\maketitle
\begin{abstract}
Quantified formulas with Uninterpreted Functions (UFs) over non-linear real arithmetic pose fundamental challenges for Satisfiability Modulo Theories (SMT) solving. Traditional quantifier instantiation methods struggle because they lack semantic understanding of UF constraints, forcing them to search through unbounded solution spaces with limited guidance.
We present AquaForte, a framework that leverages Large Language Models to provide semantic guidance for UF instantiation by generating instantiated candidates for function definitions that satisfy the constraints, thereby significantly reducing the search space and complexity for solvers. Our approach preprocesses formulas through constraint separation, uses structured prompts to extract mathematical reasoning from LLMs, and integrates the results with traditional SMT algorithms through adaptive instantiation.
AquaForte maintains soundness through systematic validation: LLM-guided instantiations yielding SAT solve the original problem, while UNSAT results generate exclusion clauses for iterative refinement. Completeness is preserved by fallback to traditional solvers augmented with learned constraints. 
Experimental evaluation on SMT-COMP benchmarks demonstrates that AquaForte solves numerous instances where state-of-the-art solvers like Z3 and CVC5 timeout, with particular effectiveness on satisfiable formulas. Our work shows that LLMs can provide valuable mathematical intuition for symbolic reasoning, establishing a new paradigm for SMT constraint solving.
\end{abstract}

\begin{links}
    \link{Code}{https://github.com/kylelv2000/Aqua-Forte}
\end{links}
\section{Introduction}

Satisfiability Modulo Theories (SMT) solving~\cite{DBLP:series/txtcs/KroeningS16,DBLP:series/faia/BarrettSST21} has become an indispensable computational foundation for numerous applications including software verification~\cite{DBLP:journals/jar/BeyerDW18}, program analysis~\cite{DBLP:conf/cav/GavrilenkoLFHM19,zhang2001constraint}, model checking~\cite{DBLP:conf/kbse/CordeiroFM09}, and automated test generation~\cite{DBLP:conf/nfm/PeleskaVL11,zhang2000specification}. Among the various SMT theories, Quantified Uninterpreted Functions with Non-linear Integer and Real Arithmetic (QUFNIRA) poses particularly formidable computational challenges. This theory naturally emerges when modeling hybrid systems that combine discrete control logic with continuous dynamics~\cite{cimatti2013smt,cimatti2015hycomp}, requiring reasoning about both integer and real-valued variables alongside abstract computational components represented as uninterpreted functions.

The undecidability of QUFNIRA stems from the interaction of quantifiers over infinite mixed integer-real domains, uninterpreted functions, and non-linear arithmetic constraints~\cite{becker2019axiom, bjorner2020navigating}. This forces practical solvers to rely on incomplete instantiation techniques, where success critically depends on generating effective quantifier instantiations that can bridge discrete and continuous reasoning.

Current state-of-the-art approaches for QUFNIRA predominantly rely on syntactic instantiation techniques, most notably E-matching~\cite{de2007efficient,reynolds2018revisiting} and model-based quantifier instantiation (MBQI)~\cite{ge2009complete}. E-matching identifies instantiation opportunities through syntactic pattern matching within the solver's congruence closure, while MBQI iteratively constructs candidate models and verifies their consistency against quantified constraints. Leading solvers such as Z3~\cite{DBLP:conf/tacas/MouraB08} and CVC5~\cite{DBLP:conf/tacas/BarbosaBBKLMMMN22} employ increasingly sophisticated heuristics to guide these processes, yet they remain fundamentally limited by their purely syntactic treatment of uninterpreted functions.

This syntactic approach creates a profound disconnect between the mathematical richness of real-world applications and the solver's reasoning capabilities. In practice, uninterpreted functions often represent well-understood mathematical concepts—distance metrics exhibiting triangle inequality properties, cost functions with monotonicity constraints, or physical transformations preserving certain invariants. However, current solvers treat these functions as completely opaque symbols, constrained only by functional congruence. This semantic blindness forces solvers to explore enormous instantiation spaces without leveraging mathematical intuition that could dramatically reduce search complexity.

The limitations of syntactic approaches manifest through two critical challenges that significantly impact solving performance. \textbf{Semantic opacity} prevents solvers from exploiting inherent mathematical properties of uninterpreted functions, causing them to miss natural simplifications and structural relationships that could lead to immediate proof discovery or efficient refutation. \textbf{Instantiation inefficiency} results from the inability to generate semantically meaningful ground terms, leading to extensive exploration of irrelevant instantiation spaces that provide little useful information for the underlying decision procedures.

To address these fundamental limitations, we propose AquaForte, a novel framework that integrates Large Language Models (LLMs) with traditional SMT solving to provide semantic guidance for QUFNIRA. Our approach leverages LLMs' extensive training on mathematical literature to analyze the contextual usage patterns of uninterpreted functions within constraint systems and generate semantically plausible concrete instantiations. Rather than treating uninterpreted functions as purely syntactic objects, our method employs LLMs to infer concrete mathematical definitions that preserve the essential behavioral properties while providing the algebraic structure necessary for efficient reasoning over mixed integer-real arithmetic constraints.

The key insight is that LLMs can identify semantic patterns in uninterpreted function usage that escape syntactic analysis, then generate concrete function instantiations preserving essential mathematical properties~\cite{frieder2023mathematical}. This semantic-guided instantiation enables solvers to work with structured mathematical expressions rather than opaque symbols, dramatically improving reasoning efficiency over mixed arithmetic domains.

Our contributions include: (1) the first systematic methodology for leveraging LLM semantic understanding to guide quantifier instantiation in QUFNIRA, transcending the limitations of purely syntactic pattern matching; (2) a soundness-preserving integration framework that provides flexible compatibility with diverse solvers, maintaining formal verification guarantees while maximizing the benefits of LLM mathematical intuition; and (3) comprehensive experimental evaluation across 1,481 benchmark instances demonstrating substantial performance improvements: 80.0\% increase in solved instances for Z3 and 183.6\% improvement for CVC5, with particularly significant gains on satisfiable cases where semantic guidance proves most impactful.

\begin{figure*}[ht] 
  \centering
  \includegraphics[width=0.9\textwidth]{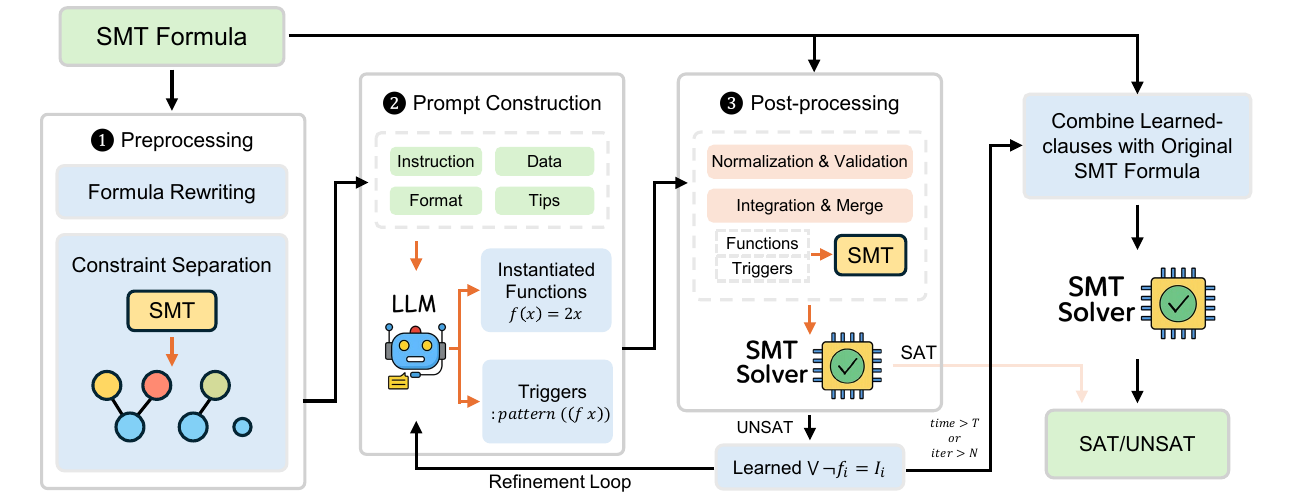}
  \caption{Overview of AquaForte.}
  \label{fig:mainalg}
\end{figure*}

\section{Preliminaries}
\label{sec:preliminaries}

We briefly review the mathematical foundations for SMT solving over uninterpreted functions with quantification, focusing on the challenging combination that motivates our LLM-guided approach.

\subsection{SMT Foundations}

Satisfiability Modulo Theories (SMT) extends propositional satisfiability by incorporating first-order theories. An SMT instance consists of a quantifier-free first-order formula $\varphi$ over a background theory $\mathcal{T}$.

\begin{definition}[SMT Problem]
Given a theory $\mathcal{T}$ and a formula $\varphi$, the SMT problem asks whether there exists a $\mathcal{T}$-model $\mathcal{M}$ such that $\mathcal{M} \models \varphi$. If such a model exists, $\varphi$ is \emph{$\mathcal{T}$-satisfiable}; otherwise, it is \emph{$\mathcal{T}$-unsatisfiable}.
\end{definition}

Modern SMT solvers such as Z3 and CVC5 employ the DPLL(T) framework \cite{nieuwenhuis2006solving}. Our work focuses on the challenging combination of real arithmetic $\mathcal{T}_{\mathbb{R}}$ and uninterpreted functions $\mathcal{T}_{UF}$:

\textbf{Real Arithmetic Theory $\mathcal{T}_{\mathbb{R}}$} enables reasoning about polynomial constraints over real numbers:
\begin{align}
\text{Terms: } \quad t &::= x \mid c \mid t_1 + t_2 \mid t_1 \cdot t_2 \mid -t \\
\text{Formulas: } \quad \varphi &::= t_1 = t_2 \mid \neg \varphi \mid \varphi_1 \land \varphi_2 \mid \varphi_1 \lor \varphi_2
\end{align}
where $x$ ranges over real variables and $c \in \mathbb{Q}$.

\textbf{Uninterpreted Functions Theory $\mathcal{T}_{UF}$} introduces function symbols constrained solely by the congruence axiom:
\begin{equation}
\forall f \in \mathcal{F}, \forall \mathbf{x}, \mathbf{y}: \bigwedge_{i=1}^n (x_i = y_i) \rightarrow f(\mathbf{x}) = f(\mathbf{y})
\end{equation}

\begin{definition}[NIRA]
The theory of Non-linear Integer and Real Arithmetic (NIRA), denoted as $\mathcal{T}_{NIRA}$, combines non-linear arithmetic constraints over integer and real domains.
\end{definition}
This theory represents one of the most challenging fragments in SMT solving.

\begin{definition}[Quantified NIRA]
When extended with quantification, NIRA formulas take the form:
\begin{equation}
\varphi ::= \text{QF-formula} \mid (\forall \vec{x}:\vec{S}.\varphi) \mid (\exists \vec{x}:\vec{S}.\varphi)
\end{equation}
where $\vec{x} = (x_1,\ldots,x_n)$ are distinct variables with sorts $\vec{S} = (S_1,\ldots,S_n)$, and each $S_i \in \{\mathbb{Z}, \mathbb{R}\}$, and QF-formula denotes the quantifier-free formulas defined previously.
\end{definition}

The computational complexity critically depends on \emph{quantifier alternation depth}. For instance, in Presburger arithmetic, formulas with bounded quantifier alternation have lower complexity than the general 2EXPTIME-complete case~\cite{fischer1998super}.

\subsection{Solving Techniques for Quantified NIRA}
Given the high computational complexity of quantified NIRA, practical SMT solvers employ \emph{quantifier instantiation} techniques to reduce quantified formulas to ground instances~\cite{reynolds2013quantifier}. For universally quantified formula $\forall x.\varphi(x)$ where $x$ ranges over mixed integer-real domains, instantiation produces:
\begin{equation}
\forall x.\varphi(x) \quad \leadsto \quad \bigwedge_{t \in T_{\mathbb{Z}} \cup T_{\mathbb{R}}} \varphi(t) \label{eq:nira_instantiation}
\end{equation}
where $T = \{t_1,\ldots,t_n\}$ is a strategically chosen set of ground terms from both integer and real domains.

\textbf{E-matching with Trigger Patterns.} The predominant instantiation approach in modern SMT solvers \cite{DBLP:conf/tacas/MouraB08} uses trigger-based pattern matching. For quantified subformula $\forall \bar{x}. \psi(\bar{x})$, a \emph{trigger pattern} is a set of terms $\{t_1(\bar{x}), \ldots, t_k(\bar{x})\}$ that appear in $\psi(\bar{x})$ and collectively mention all quantified variables. The solver instantiates the quantified formula only when ground terms matching these patterns appear during search.

\begin{example}[E-matching Process]
Given $\forall x. P(f(x)) \land f(x) > 0$ with trigger pattern $f(x)$, and ground terms $\{f(3), f(a), 5\}$ in the current context, E-matching instantiates the formula at $x = 3$ and $x = a$, producing $P(f(3)) \land f(3) > 0 \land P(f(a)) \land f(a) > 0$.
\end{example}

\textbf{Model-Based Quantifier Instantiation (MBQI).} MBQI \cite{ge2009complete} constructs candidate models for quantifier-free portions, then validates quantified constraints. When a quantified formula is violated in the candidate model, the violating assignment provides witness terms for subsequent instantiation. This approach is more semantic than pure pattern matching but remains limited by the black-box treatment of uninterpreted functions.

\subsection{Large Language Models and Prompt}

Large Language Models (LLMs) are neural networks based on the Transformer architecture \cite{vaswani2017attention}, trained on vast text corpora to predict the next token in a sequence. Modern LLMs like GPT-4 \cite{achiam2023gpt} demonstrate emergent mathematical reasoning capabilities \cite{lewkowycz2022solving,wei2022emergent} through scale and self-attention mechanisms that capture long-range dependencies, developing internal representations that encode semantic relationships and domain-specific knowledge.

\textbf{Prompt Engineering.} LLM performance critically depends on \emph{prompt design}—the systematic construction of input prompts to elicit desired behaviors~\cite{liu2023pre,sahoo2024systematic}. Effective prompts include: (1) \emph{Task specification} with clear instructions; (2) \emph{Context provision} with relevant background; (3) \emph{Output formatting} specifying desired structure; (4) \emph{Few-shot examples} demonstrating expected patterns~\cite{brown2020language}. Advanced techniques include chain-of-thought prompting \cite{wei2022chain} for step-by-step reasoning and role-based prompting to invoke domain expertise.

\begin{example}[Structured Prompt Design]
A mathematical prompt might follow: ``You are a mathematics expert. Given constraints: [problem], analyze relationships and provide: [format]. Example: [demonstration]. Requirements: [constraints]."
\end{example}

\textbf{Integration with Formal Systems.} The complementary nature of LLMs and formal systems enables hybrid architectures where LLMs provide semantic insights and hypothesis generation, while symbolic systems ensure logical correctness and verification. This synergy combines intuitive reasoning with formal precision.

\section{Methodology}
\label{sec:method}

Figure~\ref{fig:mainalg} presents \textbf{AquaForte}, a novel framework that leverages Large Language Models to provide semantic guidance for uninterpreted functions in SMT solving. This section describes our approach through four main components: an overview with illustrative examples, uninterpreted function instantiation techniques, auxiliary strategies for trigger generation, and integration with traditional SMT algorithms.

\subsection{Overview}
\label{subsec:overview}

To illustrate our approach, consider the following SMT formula containing constraints on uninterpreted functions:
\begin{equation}
\varphi = \forall x \in \mathbb{R}. \; f(2x) = 2x \land g(2x) = 2x \label{eq:scaling_constraint}
\end{equation}
This formula defines both functions $f$ and $g$ to have identical behavior at scaled inputs: $f(2x) = g(2x) = 2x$ for all $x$. From a mathematical perspective, it is intuitive that $f(x) = g(x)$ should hold universally. However, traditional SMT solvers struggle to derive this equivalence relationship and often timeout when queried about $\forall x. f(x) = g(x)$.

Traditional SMT solvers rely on axiomatization-based derivation methods, constructing proof processes through predefined inference rules and quantifier instantiation patterns. However, since the constraint only provides information about $f$ and $g$ at points of the form $2x$, the solver cannot directly establish equivalence at arbitrary points $x$. The instantiation process becomes an infinite search that fails to find contradictions, leading to timeout.

Our core insight is to leverage LLM's mathematical intuition to recognize semantic relationships between functions. Based on this semantic understanding, the LLM suggests candidate function instantiations such as $f(x) = x$ and $g(x) = x$, which satisfy the original constraints and make the equivalence $f(x) = g(x)$ immediately verifiable. We instantiate uninterpreted functions with concrete mathematical expressions via semantic analysis, thereby eliminating quantified reasoning bottlenecks and accelerating SAT solving.

\subsection{Uninterpreted Function Instantiation}
\label{subsec:instantiation}

This process comprises three stages: Preprocessing, LLM Instantiation, and Post-processing, transforming raw SMT formulas into semantically meaningful instantiations.

\subsubsection{Preprocessing}
Algorithm~\ref{alg:preprocessing} details the preprocessing stage that applies two techniques to reduce cognitive complexity while maintaining mathematical structure.

\textbf{Formula Rewriting.} We apply semantic-preserving transformations: (1) formula simplification through expression normalization, redundant operation elimination, and tautology / contradiction reduction to Boolean constants; (2) interpreted function expansion to expose mathematical semantics.

\textbf{Constraint Separation.} Using union-find structures, we identify connected components of constraints sharing uninterpreted functions. Functions co-occurring in any constraint merge into the same component, decomposing the formula into independent subproblems. This reduces individual query complexity from $O(|\mathcal{F}|^2)$ to $O(|C_i|^2)$ where $C_i$ denotes the $i$-th component, potentially achieving $|C_i| \ll |\mathcal{F}|$ when constraints are well-separated.

\begin{algorithm}[t]
\caption{Formula Rewriting and Purification}
\label{alg:preprocessing}
\begin{algorithmic}[1]
\STATE \textbf{Input:} Formula $\varphi$, uninterpreted functions $\mathcal{F} = \{f_1, \ldots, f_k\}$
\STATE \textbf{Output:} Formula components $\mathcal{C} = \{C_1, \ldots, C_m\}$
\STATE $\varphi' \leftarrow \textsc{RewriteFormula}(\varphi)$ \COMMENT{Apply simplifications and expansions}
\STATE Initialize union-find structure $UF$ with functions $\mathcal{F}$
\STATE $constraints \leftarrow \textsc{ExtractConstraints}(\varphi')$
\FOR{each constraint $c \in constraints$}
    \STATE $funcs \leftarrow \textsc{GetUninterpretedFunctions}(c)$
    \FOR{each pair $(f_i, f_j) \in funcs \times funcs$}
        \STATE $\textsc{Union}(UF, f_i, f_j)$
    \ENDFOR
\ENDFOR
\STATE Group constraints by their representatives in $UF$
\RETURN Connected components $\mathcal{C}$
\end{algorithmic}
\end{algorithm}

\subsubsection{LLM Instantiation}
For each connected component $C_i$, we use a Large Language Model to generate concrete instantiations $\mathcal{I} = \{I_1, I_2, \ldots, I_k\}$ that materialize each uninterpreted function $f_i \in \mathcal{F}$ into explicit mathematical definitions such as polynomial functions, piecewise functions, or other computable constructs.

We construct a structured prompt $\mathcal{P}_i$ as:
$$\mathcal{P}_i = \langle \mathcal{S}_{instruction}, \mathcal{D}_{data}, \mathcal{O}_{format}, \mathcal{T}_{tips} \rangle$$
where $\mathcal{S}_{instruction}$ establishes the mathematical interpretation task, $\mathcal{D}_{data}$ provides the constraints from component $C_i$ and uninterpreted function signatures, $\mathcal{O}_{format}$ defines a structured JSON output format requiring reasoning chains and confidence assessments, and $\mathcal{T}_{tips}$ specifies implementation constraints for valid SMT-LIB syntax~\cite{BarFT-RR-25}.

The prompt instructs the LLM to ``understand the mathematical meaning of the SMT formula and reason from a mathematical perspective to derive the concrete form of the uninterpreted function." This hierarchical approach ensures mathematically grounded instantiations while constraining the output space to valid SMT-LIB~\cite{barrett2010smt} constructs. The structured output format promotes transparency by requiring explicit justification for each proposed instantiation, enabling quality assessment and seamless integration with existing SMT solver infrastructure.

\subsubsection{Post-processing}
\label{subsec:postprocessing}
The LLM output undergoes a systematic validation pipeline to ensure correctness and compatibility with SMT solvers. First, we normalize the raw response by extracting the structured JSON definitions while discarding any conversational text. Each proposed instantiation is then parsed and type-checked against SMT-LIB syntax requirements; invalid expressions trigger a re-query to the LLM with corrective feedback.

Finally, instantiations from all components are merged. During this crucial step, we maintain a global symbol table to detect and resolve potential issues such as \emph{variable name conflicts} while verifying type consistency across function signatures. This systematic pipeline transforms raw LLM output into validated, syntactically correct instantiations ready for integration into the original SMT formula.

\subsection{Quantifier Elimination Strategies}
Our trigger generation approach leverages large language models to automatically synthesize instantiation patterns from quantified formula constraints. Given a quantified formula $\varphi$, we directly present the logical constraints to the LLM and task it with generating effective trigger patterns.

The LLM processes these quantified constraints and generates candidate triggers $T = \{t_1, t_2, \ldots, t_k\}$ that capture essential instantiation patterns. Each generated trigger $t_i$ represents a heuristic rule that identifies when specific quantifier instantiations are likely to be productive for proof search. The key insight is that modern LLMs possess sufficient logical reasoning capabilities to recognize structural patterns within quantified formulas and translate them into actionable instantiation strategies.

The generated triggers undergo systematic validation to ensure syntactic correctness and semantic consistency. We employ filtering mechanisms to remove malformed patterns and consolidate semantically equivalent triggers to prevent redundancy. The validated triggers are then directly integrated into the quantifier instantiation module, where they guide the selection of instantiation terms during automated proof search.

\begin{algorithm}[t]
\caption{Adaptive LLM-Guided SMT Solving}
\label{alg:adaptive_smt}
\begin{algorithmic}[1]
\STATE \textbf{Input:} Formula $\varphi$, max iterations $N$, time budget $T$
\STATE \textbf{Output:} \texttt{SAT}, \texttt{UNSAT}, or \texttt{UNKNOWN}

\STATE \textbf{Initialize:} $\mathit{history} \leftarrow \emptyset$, $\mathit{learned} \leftarrow \emptyset$, $\mathit{iter} \leftarrow 0$
\STATE $\mathit{start} \leftarrow \textsc{Time}()$

\WHILE{$\mathit{iter} < N$ \AND $\textsc{Time}() - \mathit{start} < T$}
    \STATE $(\mathcal{I}, \mathcal{T}) \leftarrow \textsc{QueryLLM}(\varphi, \mathit{history})$ \COMMENT{Get instantiations}
    \STATE $\varphi_{\mathit{inst}} \leftarrow \textsc{ApplyInstantiations}(\varphi, \mathcal{I})$
    \STATE $\varphi_{\mathit{inst}} \leftarrow \textsc{AddTriggers}(\varphi_{\mathit{inst}}, \mathcal{T})$
    
    \STATE $\mathit{result} \leftarrow \textsc{SMTSolver}(\varphi_{\mathit{inst}})$ \COMMENT{Bounded solving}
    \IF{$\mathit{result} = \texttt{SAT}$}
        \RETURN \texttt{SAT}
    \ELSIF{$\mathit{result} = \texttt{UNSAT}$}
        \STATE $\psi \leftarrow \bigvee_{i=1}^k \neg(f_i = I_i)$ \COMMENT{Exclusion clause}
        \STATE $\mathit{history} \leftarrow \mathit{history} \cup \{(\mathcal{I}, \texttt{refuted})\}$
        \STATE $\mathit{learned} \leftarrow \mathit{learned} \cup \{\psi\}$
    \ELSE
        \STATE $\mathit{history} \leftarrow \mathit{history} \cup \{(\mathcal{I}, \texttt{timeout})\}$
    \ENDIF
    \STATE $\mathit{iter} \leftarrow \mathit{iter} + 1$
\ENDWHILE

\STATE Fallback: \RETURN $\textsc{SMTSolver}(\varphi \land \bigwedge \mathit{learned})$
\end{algorithmic}
\end{algorithm}

\subsection{Integration with Traditional Algorithms}
\label{subsec:integration}

We present a hybrid framework that systematically integrates LLM guidance with traditional SMT solving through iterative refinement cycles, where failed instantiation attempts provide structured feedback for subsequent queries.

Algorithm~\ref{alg:adaptive_smt} queries the LLM with augmented information $\mathcal{Q} = \langle \varphi, \mathit{history}, \mathcal{F} \rangle$ (line 4), where $\varphi$ is the original formula, $\mathit{history}$ contains previous failures with contexts, and $\mathcal{F}$ specifies uninterpreted functions requiring instantiation. This enriched context enables the LLM to avoid unsuccessful interpretations and explore alternative mathematical relationships. The LLM response $(\mathcal{I}, \mathcal{T})$ provides instantiations and trigger patterns, systematically applied to construct the modified formula $\varphi'$ (lines 5-6).

The validation phase (lines 7-9) applies bounded SMT solving with time limit $\tau_1$. Upon satisfiability, the algorithm terminates with SAT. Otherwise, failure analysis records the failed instantiation $\mathcal{I}$ in $\mathit{history}$ (line 12) and generates exclusion clause $\psi = \bigvee_{i=1}^k \neg(f_i = I_i)$ to constrain future search (line 13).

Upon reaching iteration threshold $N$ or time budget $T$ (line 20), systematic fallback consolidates all exclusion clauses into $\varphi_{augmented} = \varphi \land \bigwedge \mathit{learned\_clauses}$. The complete SMT solver operates on this constraint-enriched formula, where learned clauses eliminate previously explored interpretation regions, often yielding improved performance over naive application.

This fallback mechanism preserves the correctness and does not weaken the base solver: when LLM-guided instantiations expire within resource limits, the procedure reduces to the traditional SMT solving over the accumulated constraints, matching the decision capability of the base solver in the worst case.

\begin{table*}[ht]
\centering
\setlength{\tabcolsep}{1mm}
\begin{tabular}{l|cc|cc|cc|cc|cc|cc}
\toprule
\multirow{2}{*}{\textbf{Method}} & \multicolumn{4}{c|}{\textbf{SMT-COMP (281 instances)}} & \multicolumn{4}{c|}{\textbf{MFD (600 instances)}} & \multicolumn{4}{c}{\textbf{SOS (600 instances)}} \\
& SAT & Time(s) & UNSAT & Time(s) & SAT & Time(s)  & UNSAT & Time(s) & SAT & Time(s) & UNSAT & Time(s) \\
\midrule
Z3 Solver & 16 & 0.15 & 104 & 0.02 & 275 & 0.08 & \textbf{29} & \textbf{0.02} & 12 & 0.02 & 0 & / \\
\textbf{AF+Z3 (GPT-4.1)} & \textbf{58} & \textbf{0.09*} & \textbf{107} & \textbf{0.10*} & \textbf{429} & \textbf{0.07*} & \textbf{29} & \textbf{0.08*} & \textbf{162} & \textbf{0.05*} & 0 & / \\
CVC5 Solver & 0 & / & 91 & 0.70 & 108 & 0.01 & 27 & 0.01 & 0 & / & 0 & / \\
AF+CVC5 (GPT-4.1) & 50 & 0.03* & 95 & 0.94* & 311 & 0.04* & 26 & 0.06* & 159 & 0.03* & 0 & / \\
\bottomrule
\end{tabular}%

\caption{Detailed performance breakdown across benchmark suites with 24-second timeout. Execution times shown exclude LLM inference latency (mean: 7.60s).}
\label{tab:detailed_results}
\end{table*}

\section{Experimental Evaluation}
\label{sec:experiments}

We conduct a comprehensive empirical evaluation to assess the effectiveness of our proposed \textbf{AquaForte (AF)} framework for LLM-guided SMT solving. 
To facilitate reproducibility, our code and benchmarks are publicly available on GitHub.
Our evaluation is designed to answer the following research questions:

\begin{itemize}
\item \textbf{RQ1:} How effective is LLM-guided instantiation compared to state-of-the-art SMT solvers?
\item \textbf{RQ2:} Does increased computational time benefit traditional solvers versus LLM-guided approaches?
\item \textbf{RQ3:} What is the effect of increasing LLM iteration counts on the number of problem instances solved?

\end{itemize}

\subsection{Benchmark Suite}
\subsubsection{SMT-COMP 2025 Benchmarks}
We evaluated our approach in the entire UFNIRA and UFLRA benchmark sets from SMT-LIB, comprising 281 instances in total. These categories combine uninterpreted functions with arithmetic constraints—UFNIRA includes non-linear integer and real arithmetic, while UFLRA focuses on linear real arithmetic.

\subsubsection{Custom Benchmark Construction}
To assess specific algorithmic capabilities beyond standard benchmarks, we developed two specialized datasets that will be released alongside our source code.

\textbf{Sum-of-Squares (SOS) Verification Dataset:} Inspired by Hilbert's 17th problem, we created 600 instances for sum-of-squares decomposition. Each instance determines whether a polynomial $F$ can be expressed as exactly three squares:
\begin{equation}
F(\mathbf{x}) = g_1^2(\mathbf{x}) + g_2^2(\mathbf{x}) + g_3^2(\mathbf{x})
\end{equation}
where $\mathbf{x} = (x_1, \ldots, x_n)$.

We employ constructive generation by creating $m$ source polynomials and computing $F = \sum_{i=1}^m p_i^2$. Problem complexity is controlled via variable dimension $n \in \{1,2,3\}$ and source count $m \in \{1,2,3,4\}$. The SMT encoding seeks uninterpreted functions $f_a, f_b, f_c$ satisfying:
\begin{align}
\forall \mathbf{x}: f_a^2(\mathbf{x}) + f_b^2(\mathbf{x}) + f_c^2(\mathbf{x}) = F(\mathbf{x})
\end{align}
Instances with $m \leq 3$ are satisfiable, while $m = 4$ cases require ``compressing" four squares into three, creating challenging algebraic reasoning problems. The dataset provides 12 parameter configurations with 50 instances each.

\textbf{Mathematical Functions Dataset (MFD):} We constructed 600 randomly generated UFNIRA instances across four categories (150 each): \textbf{Rational function inequalities} combine existential quantifiers with comparison operators over function evaluations at randomly sampled domain points ($\exists x: f(x) \neq f(0)$). \textbf{Piecewise function inequalities} partition the real domain into 2-4 intervals with randomly assigned linear expressions and inequality constraints at boundary points. \textbf{Recursive function problems} employ template instantiation of recurrence patterns with varying recursion depths ($g(x) = f(x) + g(x-1)$), combined with randomly selected initial conditions and positivity constraints. \textbf{Function limit problems} encode linear algebraic properties (additivity: $f(x+y) = f(x) + f(y)$, homogeneity: $f(kx) = k \cdot f(x)$) with automatically generated existential boundary conditions that test contradictory inequalities.

Our complete benchmark suite totals 1481 instances, spanning standard verification benchmarks and specialized mathematical reasoning problems.

\subsection{Experimental Setup}
\textbf{Hardware and Environment.} All experiments were conducted on a server with AMD EPYC 7763 64-core processor (2.45 GHz) and 512 GB RAM, ensuring consistent experimental conditions.

\textbf{SMT Solvers.} We evaluate our approach using the two most prominent state-of-the-art SMT solvers:  Z3 (v4.15.0) and
    CVC5 (v1.2.1), which represent the current mainstream solutions for quantified formula solving.

\textbf{LLMs.} We employ three widely used LLMs: GPT-4.1, DeepSeek-V3~\cite{deepseek2024deepseek}, and Claude-4-Sonnet, each configured with temperature 0.01 for deterministic responses.

\textbf{Timeout Configuration.} We primarily use a 24-second timeout, following SMT-COMP's standard for rapid evaluation. We also conduct experiments with 1200-second timeout to assess long-term scalability.

\textbf{Iteration Strategy.} Given the computational cost of LLM inference, we default to single-iteration calls (N=1) unless specified otherwise, making our approach practical for real-world deployment.

\subsection{RQ1: Overall Performance Comparison}

\begin{figure}[ht]
    \centering
    \includegraphics[width=\columnwidth]{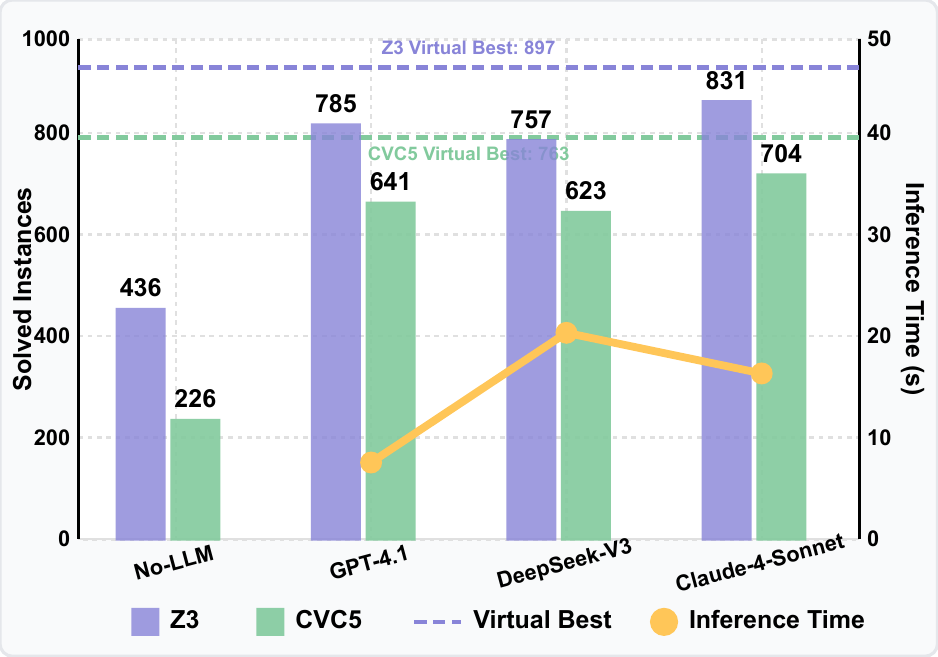}
    \caption{Total solved instances across different LLM-solver combinations with 24s timeout. Virtual Best represents the union of all LLM-enhanced configurations.}
    \label{fig:solver_comparison}
\end{figure}

Figure~\ref{fig:solver_comparison} presents the overall performance comparison across different LLM-solver combinations. The results demonstrate substantial improvements over baseline solvers: Z3 benefits from an 80.0\% improvement with GPT-4.1 (785 vs. 436 instances), while CVC5 shows an even more dramatic 183.6\% improvement (641 vs. 226 instances). Claude-4-Sonnet achieves the best single-model performance with 831 solved instances for Z3, representing a 90.6\% improvement over the baseline. The LLM inference times vary across models, with GPT-4.1 requiring 7.6 seconds, Claude-4-Sonnet 16.56 seconds, and DeepSeek-V3 20.66 seconds on average.

The Virtual Best configuration solves 897 instances with Z3 and 763 with CVC5, demonstrating complementary strengths among different LLMs. The combined approach surpasses the baselines by 105.7\% and 237.6\%, respectively.

\subsection{Detailed Benchmark Analysis}

Table~\ref{tab:detailed_results} presents a detailed breakdown of performance across individual benchmark suites. Our approach achieves particularly notable improvements on satisfiable (SAT) instances. This superior performance indicates that LLMs are highly effective at proposing appropriate instantiations that lead to satisfying assignments, leveraging their semantic understanding of underlying mathematical relationships. In addition, the approach can effectively accelerate the solver’s performance.

\subsection{Comparison of Different Synthesis Strategies}

\begin{table}[ht]
\centering
\setlength{\tabcolsep}{1mm}
\small
\begin{tabular}{l|c|c|c|c}
\toprule
\textbf{Method} & \textbf{SAT} & \textbf{UNSAT} & \textbf{Total solved} & \textbf{Improvement} \\
\midrule
Base & 108 & 118 & 226 & 0.0\% \\
MBQI & 256 & 108 & 364 & +61.1\% \\
CEGQI & 337 & 52 & 389 & +72.1\% \\
ENUM & 73 & 52 & 125 & -44.7\% \\
\textbf{GPT-4.1(Ours)} & \textbf{520} & \textbf{121} & \textbf{641} & \textbf{+183.6\%} \\
\bottomrule
\end{tabular}
\caption{Comparison of cvc5 strategies with 24s timeout.}
\label{tab:Synthesis_Strategies}
\end{table}

Recent work in function synthesis also offers several methods that can address our task. Mainstream approaches include MBQI, counterexample-guided quantifier instantiation (CEGQI)~\cite{reynolds2015counterexample}, and enumeration~\cite{reynolds2019cvc4sy} over SyGuS grammars. These methods are available in the state-of-the-art solver cvc5, so we compare our solver against the corresponding cvc5 strategies under a uniform 24s timeout per instance. As shown in Table~\ref{tab:Synthesis_Strategies}, these recent methods yield solid gains, but still fall short of our approach by a clear margin. Just as important, these techniques do not conflict with our LLM-guided design. They are complementary: LLM proposals can seed or steer CEGQI and help prune or rank enumerations, suggesting a potential path to further improve solve rates.

\subsection{RQ2: Efficiency Analysis: Rapid Solution Discovery}

\begin{table}[ht]
\centering
\small
\begin{tabular}{l|c|c|c}
\toprule
\textbf{Method (N=1)} & \textbf{24s} & \textbf{1200s} & \textbf{Improvement} \\
\midrule
Z3 Solver & 436 & 439 & +0.7\% \\
AF+Z3 (GPT-4.1) & 785 & 788 & +0.4\% \\
CVC5 Solver & 226 & 226 & +0.0\% \\
AF+CVC5 (GPT-4.1) & 641 & 641 & +0.0\% \\
\bottomrule
\end{tabular}
\caption{Performance under different timeout constraints}
\label{tab:timeout_analysis}
\end{table}

Table~\ref{tab:timeout_analysis} shows that both traditional SMT solvers and LLM-guided approaches gain little from longer timeout (24s to 1200s), indicating more time alone helps little. For traditional solvers, this reflects fundamental search strategy limitations; for LLM-guided methods, this confirms that single LLM calls either succeed rapidly or require additional iterations rather than extended runtime.

\subsection{RQ3: Multi-Iteration Analysis}

\begin{figure}[ht]
    \centering
    \includegraphics[width=\columnwidth]{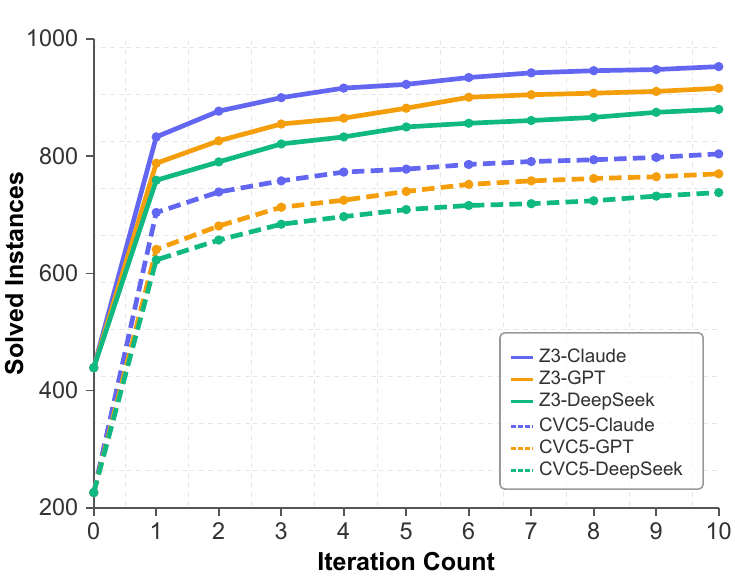}
    \caption{Performance across multiple iterations with 1200s timeout.}
    \label{fig:solver_iter_analysis}
\end{figure}

Figure~\ref{fig:solver_iter_analysis} evaluates the impact of iterative LLM guidance on solver performance. Starting from identical baselines (Z3: 439, CVC5: 226), all combinations demonstrate consistent improvements with increased iterations. Comparing single iteration (N=1) to ten iterations (N=10), Claude achieves improvements of 14.4\% for Z3 (833→953) and 14.2\% for CVC5 (704→804). On average across all LLMs, Z3 improves by 15.5\% and CVC5 by 17.6\% from iteration 1 to 10. This demonstrates that iterative refinement significantly enhances performance, with diminishing returns observed after 5-7 iterations.




\subsection{Discussion and Insights}
Our evaluation reveals where LLM-guided SMT solving excels. The approach shows striking asymmetric performance: 3.6× improvement on SAT instances versus minimal gains on UNSAT instances, indicating LLMs excel at proposing satisfying instantiations through semantic pattern recognition but struggle with exhaustive unsatisfiability proofs. This makes the approach particularly valuable for model finding tasks.
While individual LLMs exhibit varying reasoning capabilities in solving different problem instances, they demonstrate certain complementarity. The Virtual Best configuration achieves +66 instances over the best single model, indicating that ensemble approaches can effectively combine different LLM reasoning patterns.
When resources permit, multiple iterations consistently improve solving rates, though diminishing returns emerge after several iterations. Unlike traditional solvers, our approach provides substantial gains with increased computational budget.

\section{Related Work}
\label{sec:related}

Recent work has explored employing LLMs in constraint solving and logical reasoning. Logic LM~\cite{pan2023logic} combines LLMs with logical forms for enhanced logical reasoning, while others focus on problem formulation: Logic.py~\cite{kesseli2025logic} for search problems, multi-agent systems for logic puzzles~\cite{berman2024solving}, and LLM-Sym for symbolic execution~\cite{wang2024python}. However, these approaches target high-level formulation rather than instantiation mechanisms.

Traditional instantiation methods rely on syntactic heuristics, but theory-specific approaches have emerged leveraging domain properties. Notable examples include counterexample-guided instantiation for linear arithmetic~\cite{reynolds2017solving}, induction-based techniques for algebraic datatypes~\cite{reynolds2015induction}, and QSMA's model interpolation with look-ahead strategies~\cite{bonacina2023qsma}. These methods still provide limited guidance for uninterpreted functions, where semantic understanding could significantly improve instantiation effectiveness.

\section{Conclusion}
\label{sec:conclusion}

We presented AquaForte, a novel framework that uses LLMs to guide quantified SMT solving over uninterpreted functions. By converting abstract function symbols into concrete mathematical expressions, our approach addresses semantic opacity and instantiation inefficiency in traditional solvers.
Results show significant improvements on satisfiable instances but limited gains on unsatisfiable ones. Different LLMs exhibit complementary strengths suggesting potential ensemble methods. Our work demonstrates how LLMs can provide valuable semantic insights for symbolic reasoning while maintaining formal guarantees.

\nocite{ge2025sharpsmt, shi2025constraintllm, zhang2008constraint, jia2025complete}

\section{Acknowledgments}
We are grateful to the anonymous reviewers for their comments and suggestions. This work has been supported by the National Natural Science Foundation of China (NSFC) under grant No.62132020 and grant No.62572461.

\bibliography{aaai2026}

\end{document}